%% file: root.tex
\pdfoutput=1
\documentclass[letterpaper, 10 pt, journal, twoside]{IEEEtran}
\usepackage{amsmath,amsfonts}
\usepackage{array}
\usepackage{textcomp}
\usepackage{stfloats}
\usepackage{url}
\usepackage{verbatim}
\usepackage{graphicx}
\usepackage{cite}

\usepackage{graphics} 
\usepackage{epsfig} 
\usepackage{times} 
\usepackage{amssymb}
\usepackage{xcolor}
\usepackage{booktabs}
\usepackage{makecell}

\makeatletter
\let\MYcaption\@makecaption
\makeatother

\usepackage[font=footnotesize]{subcaption}

\makeatletter
\let\@makecaption\MYcaption
\makeatother

\usepackage{multirow}
\usepackage[ruled,vlined]{algorithm2e}

\makeatletter
\newcommand{\removelatexerror}{\let\@latex@error\@gobble}
\makeatother

\begin{document}

\title{Towards Accurate Loop Closure Detection in Semantic SLAM with 3D Semantic Covisibility Graphs}

\author{Zhentian Qian$^{1}$, Jie Fu$^{2}$, and Jing Xiao$^{1}$%
\thanks{Manuscript received: September, 9, 2021; Revised December, 7, 2021; Accepted January, 3, 2022.}
\thanks{This paper was recommended for publication by Editor Javier Civera upon evaluation of the Associate Editor and Reviewers' comments.} 
\thanks{$^{1}$ Zhentian Qian and Jing Xiao are with the Robotics Engineering Department, Worcester Polytechnic Institute, Worcester, Massachusetts, USA.
        {\tt\small zqian, jxiao2@wpi.edu}}%
\thanks{$^{2}$Jie Fu is with the Department of Electrical and Computer Engineering, University of Florida, Gainesville, Florida, USA.
        {\tt\small fujie@ufl.edu}}%
\thanks{Digital Object Identifier (DOI): see top of this page.}
}

\markboth{IEEE ROBOTICS AND AUTOMATION LETTERS. PREPRINT VERSION. January, 2022}%
{Qian \MakeLowercase{\textit{et al.}}: Towards Accurate Loop Closure Detection in Semantic SLAM with 3D Semantic Covisibility Graphs}


\maketitle

\begin{abstract}
\label{sec:Abstract}
\input{sections/abstract}
\end{abstract}


\section{INTRODUCTION}
\label{sec:Introduction}
\input{sections/introduction}


\section{SYSTEM OVERVIEW}
\label{sec:System Overview}
\input{sections/system_overview}

\section{OBJECT COVISIBILITY GRAPH}
\label{sec:Covisibility Graph}
\input{sections/covisibilitygraph}

\section{SEMANTIC LOOP DETECTION}
\label{sec:Loop Graph}
\input{sections/loopclousrealgorithm}

\section{EXPERIMENTAL EVALUATION}
\label{sec:Experimental Evaluation}
\input{sections/experiments}

\section{CONCLUSIONS}
\label{sec:Conclusion}
\input{sections/conclusion}
 
%
\bibliographystyle{IEEEtran}
\bibliography{bibtex/mybibfile}


 





\end{document}

%% file: sections/abstract.tex
Loop closure is necessary for correcting errors accumulated in simultaneous localization and mapping (SLAM) in unknown environments. However, conventional loop closure methods based on low-level geometric or image features may cause high ambiguity by not distinguishing similar scenarios. Thus, incorrect loop closures can occur. Though semantic 2D image information is considered in some literature to detect loop closures, there is little work that compares 3D scenes as an integral part of a semantic SLAM system. 
This paper introduces an approach, called SmSLAM+LCD, integrated into a semantic SLAM system to combine high-level 3D semantic information and low-level feature information to conduct accurate loop closure detection and effective drift reduction.
The effectiveness of our approach 
is demonstrated in testing results.  

%% file: sections/introduction.tex
\IEEEPARstart{L}{oop} closure is an essential part of a SLAM system. During a long time motion of a robot, errors often accumulate in the estimated robot poses from SLAM and cause the estimated robot trajectory to drift. Loop closure methods are used to detect 
whether the robot has returned to a place it has visited before (place recognition) and correct the accumulated localization error if it has. 
Classic loop closure methods such as the method employed in ORB-SLAM2 \cite{mur2017orb} rely on geometric features to detect loop closure and compensate for the drifting errors. Such methods can have difficulties distinguishing similar scenarios or environments with symmetry if the low-level geometric features in those environments are very similar. More recently, some researchers have considered using semantic information in SLAM to aid the task of loop closure. In the following, we provide an overview of related approaches.
\subsection{Related Work}
\input{sections/relatedwork}
\subsection{Contributions}

In this paper, we introduce a novel method for loop detection and drift correction in a semantic SLAM system based on monocular vision to take full advantage of both high-level semantic and low-level geometric information. We call our method SmSLAM+LCD (for Semantic SLAM and Loop Closure Detection). The main contributions are:
\begin{itemize}
    \item A novel approach to organize semantic object information into 3D semantic covisibility graphs. 
    \item A hierarchical loop detection approach. Loop closure candidates are first proposed based on low-level geometric features and then checked by comparing the corresponding 3D semantic covisibility subgraphs to avoid false positives. 
    \item A coarse-to-fine approach to compute the SIM(3) transformation between the candidates for loop closure. 
    \item A virtual dataset and a real-world dataset for testing if loop closure algorithms will produce false positives (available at \url{https://users.wpi.edu/%7Ezqian/#datasets}). 
    \item Testing results to demonstrate the effectiveness of the introduced SmSLAM+LCD approach. 
\end{itemize}

%% file: sections/relatedwork.tex
\subsubsection{Semantic SLAM}
Semantic SLAM methods are focused on representing, mapping, and localizing 3D objects. The pioneering work of SLAM++  \cite{salas2013slam++} builds object meshes using prior object models and an RGB-D camera. Improving on SLAM++, the work in \cite{galvez2016real} relies solely on monocular input and learns the scale of the map from prior object models. Lifting the requirement of any prior object models, QuadricSLAM \cite{nicholson2018quadricslam} and CubeSLAM \cite{yang2019cubeslam} build simplified object representations as quadric and cuboid, respectively. However, neither QuadricSLAM nor CubeSLAM addressed the issue of loop closure. In \cite{zhang2018semantic}, an octomap is used to represent objects based on semantically augmented 3D point clouds from RGB-D SLAM.

\subsubsection{Semantic loop detection}
For loop detection, one line of research seeks to extract semantic information in an image to construct an image descriptor, which is later used to compare and find images from the same scene (i.e., loop detection). The semantic information can be encoded in a vector, as done in Yang \textit{et al.} \cite{yang2018toward}, Hu \textit{et al.} \cite{hu2019loop} and Merill and Huang \cite{merrill2019calc2}. The differences among those approaches lie in how the semantic vector is constructed. For example, the semantic vector generated in \cite{yang2018toward} encodes the object class information presented in the image, while the semantic vector created in \cite{hu2019loop} encodes the object class, confidence score, and bounding box shape. On the other hand, Merill and Huang \cite{merrill2019calc2} resort to a deep learning method and construct a multi-decoder variational autoencoder (VAE) that encodes the visual appearance and semantic information to construct a global image descriptor.

Semantic information is also used to trim away the features on dynamic objects or non-discriminative features. For example, Chen \textit{et al.} \cite{chen2020semantic} mask out dynamic objects in loop closure candidate images, removing the interference from the instance-level semantic inconsistencies. The remaining features on candidate images are then used for geometric verification \cite{galvez2012bags}. Using semantic segmentation, Mousavian \textit{et al.}  \cite{mousavian2015semantically} only convert features of artificial structures to Bag of Words (BoW) vectors and discard non-discriminative features on vegetation, sky, and road. The generated BoW vectors are used for image comparison and proven to improve the accuracy of location recognition.

 Semantic information can also be encoded in a graph. Such a graph represents the observed environment, with semantic landmarks as the vertices and the edges retaining the structure information between semantic landmarks. Wang \textit{et al.} \cite{wang2020robust} and Cascianelli \textit{et al.}  \cite{cascianelli2017robust} both introduced covisibility graphs for this purpose. To detect loop closures, the overall similarity of a query image and a candidate image is composed of two components: the similarity of landmark vertices and the similarity of the neighboring information of the vertices in each image subgraph. CNN features and landmark region shapes are used to compare landmarks in both papers. While \cite{wang2020robust} uses random walk descriptors to compare the neighboring information of the nodes in each image subgraph, \cite{cascianelli2017robust} opts to use adjacency matrix. 
 
 However, those methods treat loop detection separately from a running semantic SLAM system: loop detection is not integrated into the SLAM system and does not utilize the multi-modal information generated by the semantic SLAM system. The loop detection methods only use the information presented in image frames.
 
\subsubsection{Semantic Loop Closure Integrated in Semantic SLAM} 

Only a few methods integrate semantic loop closure with semantic SLAM. For example, Liu \textit{et al.} \cite{liu2019global} extract a global semantic graph from the dense semantic map built by semantic SLAM based on RGB-D sensing and then use random walk descriptors to match the nodes in the query graph with the nodes in the global semantic graph. Benefiting from the 3D information stored in the object point clouds, the matching is only conducted coarsely. Even with inexact matching, the final localization can still be performed by aligning the semantic points associated with the matched nodes.
Li \textit{et al.} \cite{li2020view} use semantic SLAM based on monocular vision to construct cuboid semantic landmarks for objects in the environment. A similarity transformation is computed for two sets of semantic landmarks presented in the candidate and query images, respectively. However, correspondences are not calculated between the two sets of semantic landmarks, and the method exhaustively iterates over all possible matches, which can be time-consuming. Moreover, the method does not consider the structural information of objects in a scene. 

%% file: sections/system_overview.tex
\begin{figure}[!t]
  \centering
  \includegraphics[width= 0.9\linewidth]{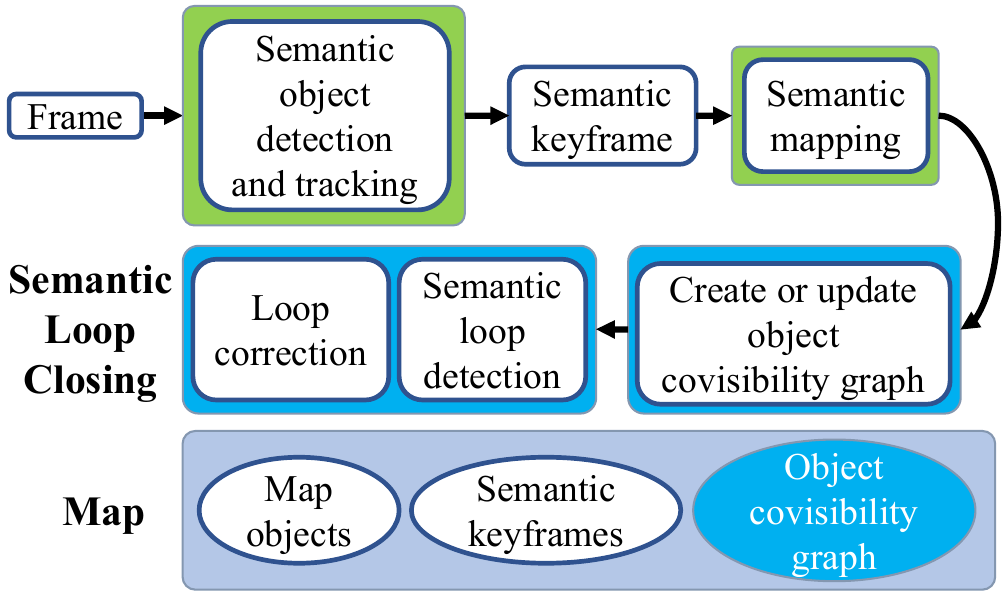}
  \caption{The SmSLAM+LCD system overview.}
  \label{fig:System Overview}
\end{figure}

We embed our approach for loop closure in a semantic SLAM system that we developed in  \cite{qian2020semantic}, featuring semantic object detection, tracking, and mapping.
Fig. \ref{fig:System Overview} provides an overview of the system. We highlight the contributions of this paper in blue.


First, we briefly review the semantic object detection, tracking, and mapping components \cite{qian2020semantic}. 
The semantic object detection and tracking component processes the incoming frames from a monocular camera. The SLAM tracking module from ORB-SLAM2 \cite{mur2017orb} is used to provide visual odometry. At the same time, an object detector (YOLOv3 \cite{redmon2018yolov3}) is used to detect the objects on the image. ORB features are then extracted in the region of interest (ROI) where objects are detected and subsequently converted to Bag of Words (BoW) \cite{galvez2012bags} vectors to describe the detected objects. The final step is to perform the object-level data association to match the detected objects to the map objects.

The semantic mapping component is performed on every new semantic keyframe $K_c$ generated by the semantic object detection and tracking component. When a new $K_c$ comes in, we update the map database to include this keyframe and the relations between map objects and semantic keyframes.

The mapped semantic objects from each image keyframe are further organized in an object covisibility graph as detailed in Section \ref{sec:Covisibility Graph}. The covisibility graph is also updated based on new observations made. 

 After the semantic mapping component processes the keyframe $K_c$, the semantic loop detection component then compares it with the keyframes stored in the map database for possible loop closures, as detailed in Section \ref{Sec:Loop Closing}. 

%% file: sections/covisibilitygraph.tex
\subsection{Graph Definition}
\label{Sec:Covisibility Graph}
An object covisibility graph $G = (V, E)$ is maintained in the map to provide a structured environment representation. 
\subsubsection{Vertices $V$}
\label{subsec:Node definition}
The vertices of the covisibility graph are the 3D semantic landmarks built by our semantic SLAM algorithm, that is, the map objects in the map database. Each mapped object is associated with a probability distribution of the object class labels to capture perception uncertainty. 
The term ``map object" or ``semantic landmark" will be used interchangeably with ``vertex" of the graph in this paper. Each map object contains the following information:
\begin{itemize}
  \item The length of the principal axes of an ellipsoid that can tightly enclose the map object, denoted by $l_a, l_b, l_c$, with $l_a$ being the length of the major axis.
  \item The position of the map object center $t$.
  \item Let $\mathcal{L}$ be the set of all object classes and $\mathcal{D}(\mathcal{L})$ be the set of probability distributions over the set $\mathcal{L}$. The vertex stores the probability distribution of the object class $p\in \mathcal{D}(\mathcal{L})$. Given an object class $z\in \mathcal{L}$, the probability of the vertex belonging to $z$ is $p( z)$.
  \item The set of keyframes sharing observations of the map object.
  \item A set of BoW vectors calculated from the image patches containing the map object on the keyframes.
\end{itemize}

\subsubsection{Edges}
An edge is added between two vertices whenever the map objects associated with the two vertices are observed in the same semantic keyframes at least three times.

\subsection{Graph Construction and Maintenance}
\begin{figure}[!t]
  \centering
  \includegraphics[width= 0.7\linewidth]{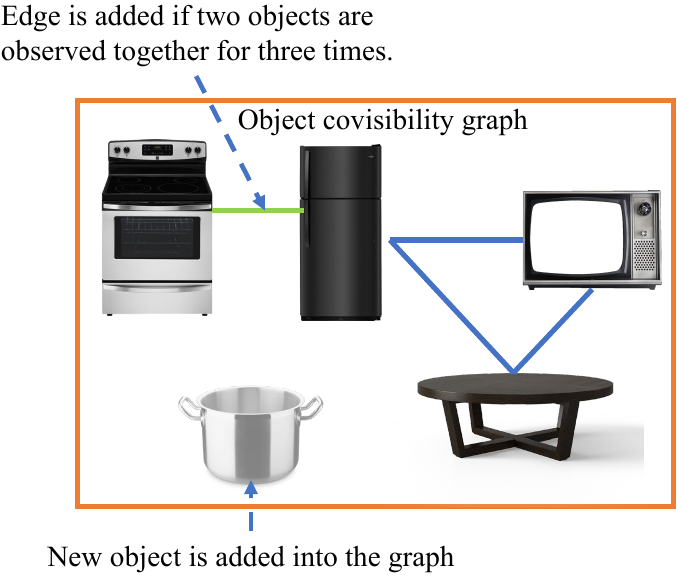}
  \caption{Updating an example object covisibility graph: a new edge and a new object are added.}
  \label{fig:graph update}
\end{figure}

The object covisibility graph is updated when a new semantic keyframe $K_c$ is created, as shown in Fig. \ref{fig:graph update}.

First, the map objects of the covisibility graph are updated. If a new object is observed in the semantic keyframe $K_c$ and is not included in the covisibility graph, such as the pot in Fig. \ref{fig:graph update}, it is added into the set of vertices. Second, the edges of the covisibility graph are updated. If an arbitrary pair of map objects $v_i$ and $v_j$ have been observed in the same semantic keyframes three times, such as the fridge and washing machine in Fig. \ref{fig:graph update}, an edge is added between them.

%% file: sections/loopclousrealgorithm.tex
\begin{figure}[!t]
  \centering
  \includegraphics[width= \linewidth]{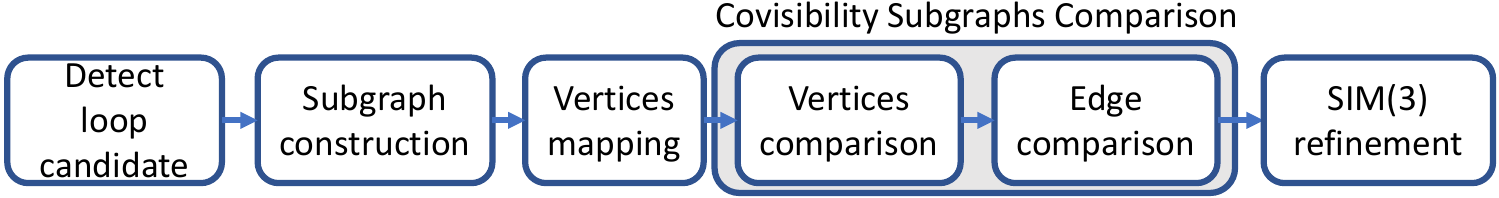}
  \caption{The semantic loop detection method.}
  \label{fig:loop clousure pipeline}
\end{figure}

\label{Sec:Loop Closing}
Fig. \ref{fig:loop clousure pipeline} outlines our method for semantic loop detection.

\subsection{Detection of Loop Candidates}
First, a set of loop closure candidates $\mathcal{K}_l$ are selected from the keyframes in the map. For the current keyframe $K_c$, keyframe $K_l \in \mathcal{K}_l$ is considered as a loop closure candidate only if the similarity score between $K_c$ and $K_l$ passes a calculated minimum threshold $s_{\text{min}}$. The similarity score is obtained by computing the $L_1$-score between the two BoW vectors ${\bf d}_c$ and ${\bf d}_l$ of frame $K_c$ and $K_l$:
\begin{equation}
    s(\mathbf{d}_c, \mathbf{d}_l) = 1 - 0.5\big\lvert \mathbf{d}_c/ \lvert \mathbf{d}_c \rvert -\mathbf{d}_l/\lvert \mathbf{d}_l\rvert \big\rvert.
    \label{eq:L1 score}
\end{equation}

We also impose the same temporal consistency in ORB-SLAM2 \cite{mur2017orb} on generating loop closure candidates.
\subsection{Subgraph Construction}
For each loop candidate $K_l$, our method further validates whether $K_l$ and $K_c$ correspond to the same place by comparing the corresponding covisibility subgraphs. Those covisibility subgraphs are vertex-induced subgraphs of the object covisibility graph stored in the map. Their vertices are only a subset of the map objects in the object covisibility graph, and their edges are directly inherited from the object covisibility graph. The vertices of the covisibility subgraph for a keyframe are the map objects observed on the keyframe; hence only a subset of the map objects are in the covisibility subgraph. 
We denote the subgraph of $K_c$ as $G_c = (V_c, E_c)$ and subgraph of $K_l$ as  $G_l = (V_l, E_l)$.

\subsection{Mapping of Vertices}

To compare two covisibility subgraphs, we need to establish the correspondence between the map objects of the two subgraphs. Benefiting from the rich information stored in each map object (refer to Section \ref{sec:Covisibility Graph}), we can obtain the correspondence by solving a maximum weighted bipartite matching problem.  For every possible map object pair $v_i \in V_c$ and $v_j \in V_l$, a similarity score is calculated:  
\begin{equation}
    s_n(i, j) =  s_a(i, j)\cdot s_c(i, j)
\end{equation}
where $s_a(i, j)$ and $s_c(i, j)$ are the appearance similarity and class similarity scores between $v_i$ and $v_j$, respectively. Empirical study suggests that using a product instead of a weighted sum between the two scores $s_a(i, j)$ and $s_c(i,j)$ yield better results. Let $d_i$ be the BoW vector converted from the image patch containing vertex $v_i$ on keyframe $K_c$, and $d_j$ be the BoW vector converted from the image patch containing vertex $v_j$ on keyframe $K_l$. The appearance similarity score $s_a(i, j)$ is computed based on the $L_1$-score between $d_i$ and $d_j$, and the formula is the same as \eqref{eq:L1 score}.
The object class similarity $s_c(i, j)$  between $v_i$ and $v_j$ employs the Bhattacharyya distance \cite{bhattacharyya1943measure}, which quantifies the distance between two class label probability distributions:
\begin{equation}
    s_c(i, j) = \sum_{z \in \mathcal{\mathcal{L}}}\sqrt{p_{i}(z)p_j(z)},
\end{equation}
where $p_{i}(\cdot)$ and $p_{j}(\cdot)$ are probability mass functions stored in vertices $i$ and $j$, as described in Section \ref{sec:Covisibility Graph}.

Next, we introduce a set of Boolean decision variables. For every possible vertex pair $v_i \in V_c$ and $v_j \in V_l$, let
\begin{equation}
    x(i, j) = \begin{cases}
    1, & \text{if } v_{i} \text{ is mapped to } v_j,\\
    0, & \text{otherwise.}
    \end{cases}
\end{equation}
$x(i, j) = 1$ represents that $ v_{i}$ and $v_{j}$ correspond to the same map object and $x(i, j) = 0$ indicates otherwise. 
The maximum weighted bipartite matching problem is then formulated as:
\begin{align}
\label{eqn:AP}
    & \max_{x(i,j) \in \{0, 1\}} \ TS=\textstyle \sum_{i=1}^{\lvert V_c \rvert} \sum_{j=1}^{\lvert V_l \rvert}s_n(i, j)x(i, j),  \\
    &\textrm{s.t.} \textstyle \sum_{i=1}^{\lvert V_c \rvert} x(i, j)  \leq 1\text{, }\forall j \text{;\quad} \sum_{j=1}^{\lvert V_l \rvert} x(i, j)  \leq 1\text{, }\forall i.
    \label{eq:constraint} 
\end{align}
The constraints \eqref{eq:constraint} mean that a vertex $v_i \in V_c$ can only be mapped to at most one vertex $v_j \in V_l$ and vice versa. 

The problem defined in (\ref{eqn:AP}) can be solved using a cost-scaling push-relabel algorithm \cite{kennedy1995solving}, readily implemented in \cite{ortools}. This algorithm has a time complexity of $\mathcal{O}(\sqrt{n}m\log{(nW)})$, where $n$ and $m$ are the number of nodes and edges in the bipartite graph, $W$ is the highest edge weight (all edge weights need to be converted to integers). 

 The map objects mapping solved in this fashion focuses on finding a mapping to maximize the total similarity score $TS$ defined in \eqref{eqn:AP}. We calculate and threshold an average similarity score $AS$ based on $TS$:
 \begin{equation}
     AS = TS/ (\textstyle \sum_{i=1}^{\lvert V_c \rvert} \sum_{j=1}^{\lvert V_l \rvert} x(i, j)) > \tau_{as}.
     \label{eq: AS}
 \end{equation}
 Loop candidates with an average similarity score $AS$ lower than the threshold $\tau_{as}$ is discarded.  A low average similarity score shows that the objects presented on the two loop candidate frames vastly differ, indicating two different places. 
 
To eliminate possible false matches, we enforce an additional post-processing step. A match $x(i, j)$ will only be accepted if the similarity score $s_n(i, j)$ is greater than $\tau_n$. Otherwise, if $s_n(i,j) < \tau_n$, $x(i,j)$ is rejected. 
 This is a practice also commonly seen in matching methods for stereo vision \cite{banks2001quantitative}. All procedures are summarized in Algorithm \ref{alg: vertices mapping}.

\begin{figure}[!hbt] 
\removelatexerror
\begin{algorithm}[H]
\caption{Mapping between Map Objects}
\SetAlgoLined
\KwIn{map object sets $V_c$ and $V_l$.}
\KwOut{all correspondence between map objects in the two sets $V_c$ and $V_l$.}
\ForEach{$v_i \in V_c$}{
\ForEach{$v_j \in V_l$}{
calculate the appearance similarity $s_a(i,j)$\;
calculate the object class similarity $s_c(i,j)$\;
calculate the similarity score $s_n(i,j)$\;
}
}
compute all $x(i,j)$ by solving the maximum weighted bipartite matching problem\;
calculate the average similarity score $AS$ using \eqref{eq: AS}\;
\lIf{$AS < \tau_{as}$}{\KwRet{NULL}}
\SetAlgoNoEnd
\ForEach{$v_i \in V_c$}{
\ForEach{$v_j \in V_l$}{
\lIf{$s_n(i, j) < \tau_{n}$}{$x(i,j) = 0$}
}
}
\KwRet{all $x(i,j)$}.
\label{alg: vertices mapping}
\end{algorithm}
\end{figure}

\subsection{Comparison of Vertices}
\label{Sec:vertices comp}
With the mapping between map objects established, we proceed to compare the map objects in $G_l$ and $G_c$. The algorithm for comparing map objects is summarized in Algorithm \ref{alg: vertices comparison}. Note that drifts are accumulated in monocular SLAM and reflected in the pose and shape information stored in the map objects of the covisibility subgraph. To account for the drifts, Algorithm \ref{alg: vertices comparison} calculates a similarity transformation from the current keyframe $K_c$ to loop keyframe $K_l$.  If the loop candidate frames $K_l$ and $K_c$ pass the map-object comparison test, a coarse SIM(3) transformation will be returned. If nothing is returned, the loop candidate is rejected. 

\begin{figure}[!hbt] 
\removelatexerror
\begin{algorithm}[H]
\SetAlgoLined
\KwIn{a set of matched vertex pairs}
\KwOut{SIM(3) transformation $S_c$}
\lIf{$N < 3$}{\KwRet{NULL}}
$N_{iter} = 0$\;
\While{$N_{iter} < $ maximum number of iterations}
{
randomly sample 3 pairs of matched map objects\;
find SIM(3) $S_c$ using the method of Horn\;
\tcp{Calculate the number of inliers}
$N_M = 0$\;
\ForEach{matched vertex pair $v_i$ and $v_j$}{
transform into common reference frame\;
\uIf{reprojection error $<$ max error  {\bf and} $\lvert l_{ai} - l_{aj} \rvert /\max (l_{ai}, l_{aj}) < \tau_s$}{
$N_M = N_M +1$
}
}
$\epsilon = N_M / N$\;
$N_{iter} = N_{iter} +1$\;
\lIf{$N_M > m_{inl}$ {\bf and} $\epsilon > \tau_{\epsilon}$}{\KwRet{$S_c$}}
}
\KwRet{NULL}.
\caption{Comparison of Map Objects}
\label{alg: vertices comparison}
\end{algorithm}
\end{figure}

A SIM(3) transformation has 7 degrees of freedom (DoF), and each pair of matched map objects provide 3 constraint equations. To compute this SIM(3) transformation, we must have a minimum of 3 pairs of matched map objects: $N = \textstyle \sum_{i=1}^{\lvert V_c \rvert} \sum_{j=1}^{\lvert V_l \rvert} x(i, j) \geq 3$. If the loop candidate frames $K_c$ and $K_l$ contain fewer than 3 pairs of matched map objects, they will be rejected. 

Taking the position information from paired map objects in $G_l$ and $G_c$, we use the RANdom SAmple Consensus (RANSAC) algorithm\cite{fischler1981random} to find a similarity transformation. Three pairs of matched map objects are randomly selected to compute a similarity transformation using the method of Horn \cite{horn1987closed}. We then check the number of inliers supported by the computed transformation. For a pair of map objects $v_i \in V_c$ and $v_j \in V_l$ to be considered inliers, they must go through the following checks. First, we check the reprojection errors \cite{hartley2003multiple} of the map object centers under the current similarity transformation. Next, we check the scale consistency of $v_i$ and $v_j$, that is, whether the two map objects are roughly of the same size. The two map objects are first transformed into a common reference frame, and their size is scaled accordingly to facilitate comparison. For scale consistency, we require: $\lvert l_{ai} - l_{aj} \rvert /\max (l_{ai}, l_{aj}) < \tau_s$, with $l_{ai}$ and $l_{aj}$ being the length of the major axis of vertex $v_i$ and $v_j$ in the common reference frame.

If the absolute number of inliers $N_M$ and relative percentage of inliers $\epsilon = N_M/N$ surpasses the preset thresholds, the computed SIM(3) transformation is accepted, and the iteration is terminated. Otherwise, another set of three matched map objects would be randomly selected. If the maximum number of iterations is reached and the similarity transformation is still not computed, the proposed loop candidate fails the vertex comparison test and is rejected. 

Note that we do not consider the orientation information or use scale information of the map objects directly in this step as the information contains some ambiguities. For any object with a similar size in two or more dimensions, its orientation cannot be uniquely defined. Multiple quadric representations with varied sizes may exist for the same object.

\subsection{Comparison of Edges}
Next, we compare the edges of the co-visibility subgraphs $G_l$ and $G_c$. Adjacency matrices are constructed for map objects in $G_l$ and $G_c$ that are matched previously, denoted as $A_l$ and $A_c$. With matches obtained from Section~\ref{Sec:vertices comp}, we can make the row and column indices of adjacency matrices $A_l$ and $A_c$ the same, i.e., make the subgraphs aligned \cite{feizi2019spectral}. The similarity between the two adjacency matrices can be measured by normalized cross-correlation \cite{stumm2015location}:
\begin{equation}
    s_e = \frac{\sum_{i=1}^{\lvert V_c \rvert} \sum_{j=1}^{\lvert V_l \rvert} A_l(i ,j)A_c(i ,j)}{\sqrt{\sum_{i=1}^{\lvert V_c \rvert} \sum_{j=1}^{\lvert V_l \rvert} A_l^2(i ,j)\sum_{i=1}^{\lvert V_c \rvert} \sum_{j=1}^{\lvert V_l \rvert} A^2_c(i ,j)}}.
\end{equation}

The edge comparison succeeds if $s_e > \tau_e$, where $\tau_e$ is a pre-defined threshold. The loop with keyframe $K_l$ is only accepted if comparisons of both edges and vertices succeed.

\subsection{SIM(3) Refinement}
As the positions of objects cannot be strictly defined, they are not precise;  Hence the SIM(3) transformation $S_c$ calculated in Algorithm \ref{alg: vertices comparison} from map object position is still coarse, as described in Section \ref{Sec:vertices comp}. To address this issue, we adopt a coarse-to-fine method to calculate the SIM(3) transformation. In the final step of our semantic loop detection method, we refine the similarity transformation using low-level map points. First, we find a set of matched map points in the two loop candidate frames $K_c$ and $K_l$. This search is sped up using the BoW vocabulary tree \cite{mur2017orb}. Next, using the coarse SIM(3) transformation $S_c$, we can enlarge the matched map points set by performing a guided search. The map points observed on frame $K_l$ are reprojected on $K_c$ based on $S_c$ to find more matches and vice versa. Again, we perform RANSAC iterations on the matched map points to arrive at a new SIM(3) transformation $S_f$. This SIM(3) transformation $S_f$ is then used to correct the drift errors.

%% file: sections/experiments.tex
We have tested our loop closure algorithms in three datasets: TUM RGB-D dataset, virtual dataset, and custom real-world dataset. We have compared our method against the loop detection method used in ORB-SLAM2 \cite{mur2017orb} and ORB-SLAM3 \cite{campos2021orb}. The parameter values used in our method
are presented in Table \ref{tab: parameter}.

\begin{table}[htbp]
\centering
\caption{Parameters and their values}
\begin{tabular}{c c c c c c} 
\toprule
$\tau_{as}$  & $\tau_{n}$ &  $\tau_{s}$ & $m_{inl}$ & $\tau_{\epsilon}$ & $\tau_e$\\ 
\midrule
0.3  & 0.008 & 0.5 & 3 & 0.59 & 0.5 \\
\bottomrule
\end{tabular}
\label{tab: parameter}
\end{table}
\subsection{Results from TUM Dataset}
For the TUM RGB-D dataset, we focus on the fr2/desk and fr3/long\textunderscore office\textunderscore household sequences, which contain loop closures and have a certain number of objects in the scene. Since sequences in the TUM dataset are not recorded in similar places, for this test, we focus on the ability of our proposed method to detect loop closures and correct drifts.   
\subsubsection{Detection Results}
To test the ability for loop detection, we wrote a script to label the loop closures in the two sequences automatically. For two keyframes to be labeled as a loop closure, their positional difference needs to be smaller than 1m, and their viewing angular difference is smaller than $53^{\circ}$. To avoid mislabeling adjacent frames as loop closures, we require the ID difference between loop closure frames to be greater than 1,000. Table \ref{tab: Precison And Recall} shows the detection results from the TUM sequences, including five-time average of the precision and recall metrics. The results suggest that when comparing with ORB-SLAM2 and ORB-SLAM3, our method detects fewer loop closure candidates and hence yields fewer true positives because of stricter criteria, but is able to reach $100\%$ precision in both sequences. 

\begin{table}[htbp]
\centering
\caption{Detection results on TUM sequences}
\begin{tabular}{c c c c c} 
\toprule
DataSet & Metrics & \textbf{Ours} & ORB-SLAM2 & ORB-SLAM3\\
\midrule
\multirow{5}{4.1em}{fr2/desk} & Detections & \textbf{23} & 28 & 297\\
& Loop Closure & \textbf{2122} & 2104 & 3067\\
& True Positives  & \textbf{23} & 28 & 51\\
& Avg. Precision  & \textbf{100\%} & 100\% & 17\%\\
& Avg. Recall & \textbf{1.1\%} & 1.4\% & 1.7\%\\
\midrule
\multirow{5}{4.1em}{\makecell{fr3/long\textunderscore\\ office\textunderscore\\ household}} & Detections  & \textbf{6}  & 37 &348\\
& Loop Closure  & \textbf{2429} & 1322 & 2529\\ 
& True Positives  & \textbf{6} & 30 & 43\\
& Avg. Precision  & \textbf{100\%} & 82\% & 13\%\\
& Avg. Recall & \textbf{0.3\%} & 2.4\% & 1.7 \% \\
\bottomrule
\end{tabular}
\label{tab: Precison And Recall}
\end{table}
\begin{figure}[!t]
     \centering
     \begin{subfigure}[b]{0.493\linewidth}
         \centering
         \includegraphics[width=1\linewidth]{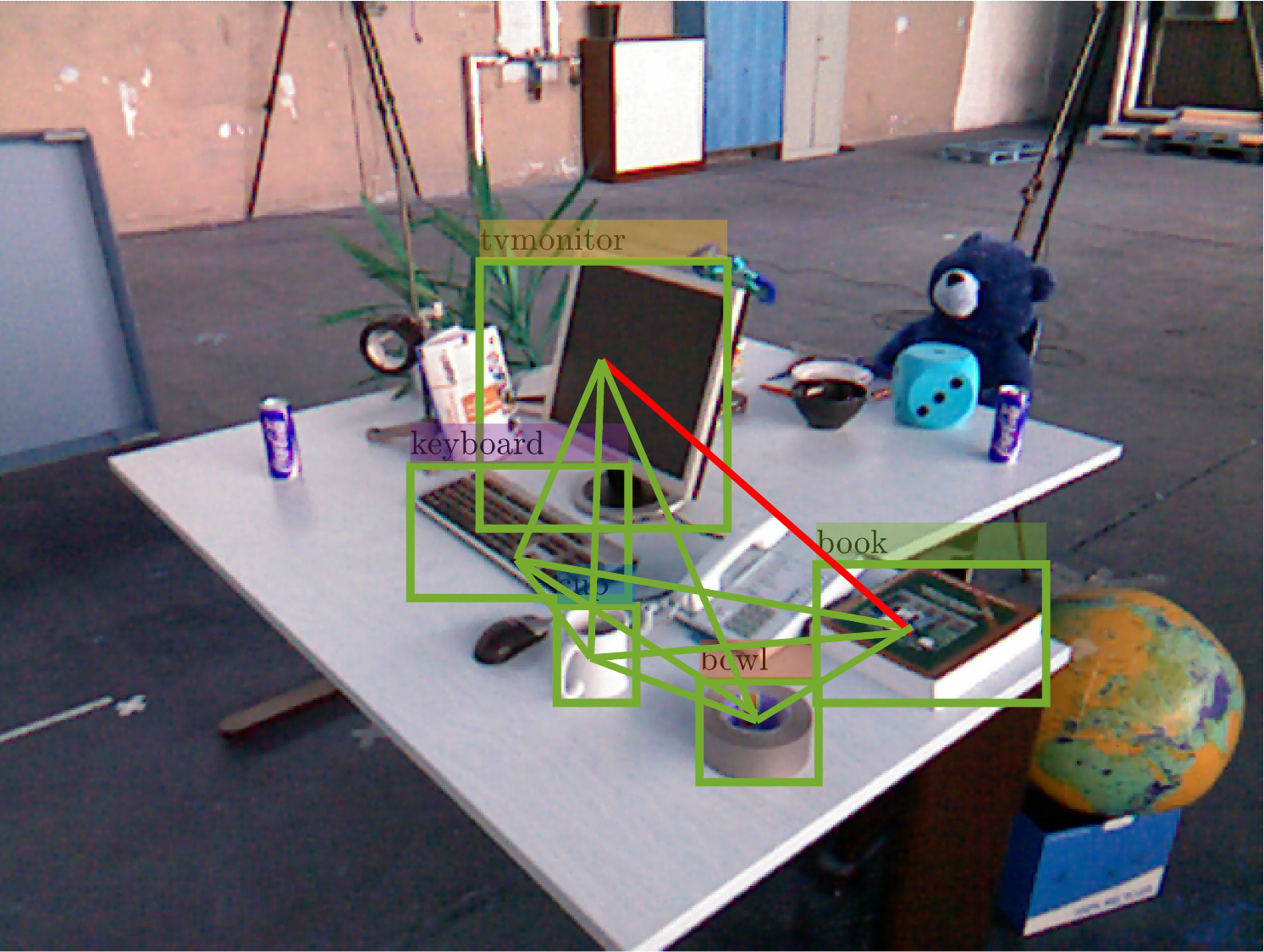}
     \end{subfigure}
     \hfill
     \centering
     \begin{subfigure}[b]{0.493\linewidth}
         \centering
         \includegraphics[width=1\linewidth]{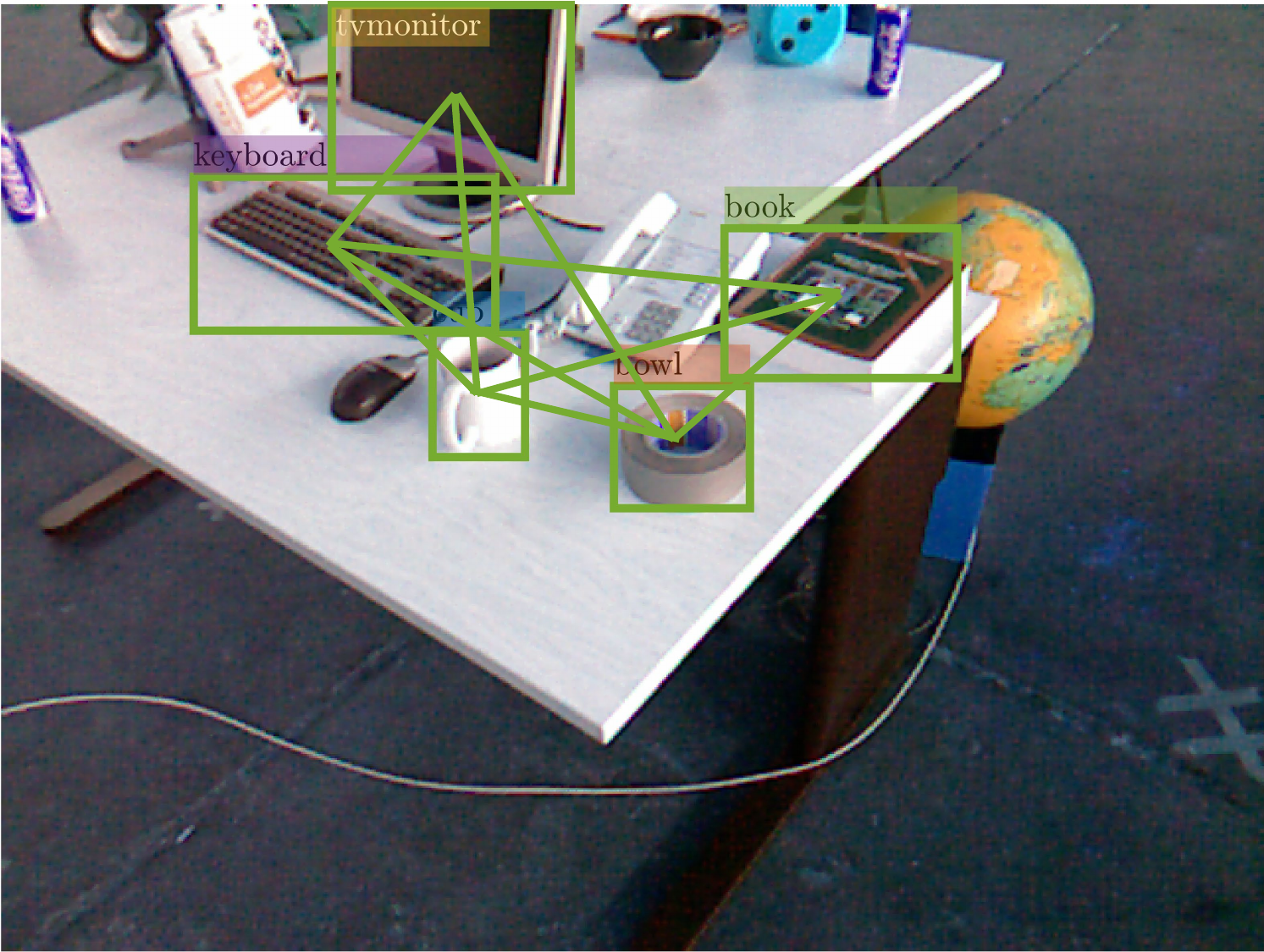}
     \end{subfigure}
        \caption{Covisibility subgraphs on two true loop closure frames (left and right) in fr2/desk sequence. Green and red plots represent matched and mismatched map objects and edges, respectively.} 
        \label{fig:graph in fr3}
\end{figure}

\begin{figure}[!t]
     \centering
     \begin{subfigure}[b]{0.493\linewidth}
         \centering
         \includegraphics[width=1\linewidth]{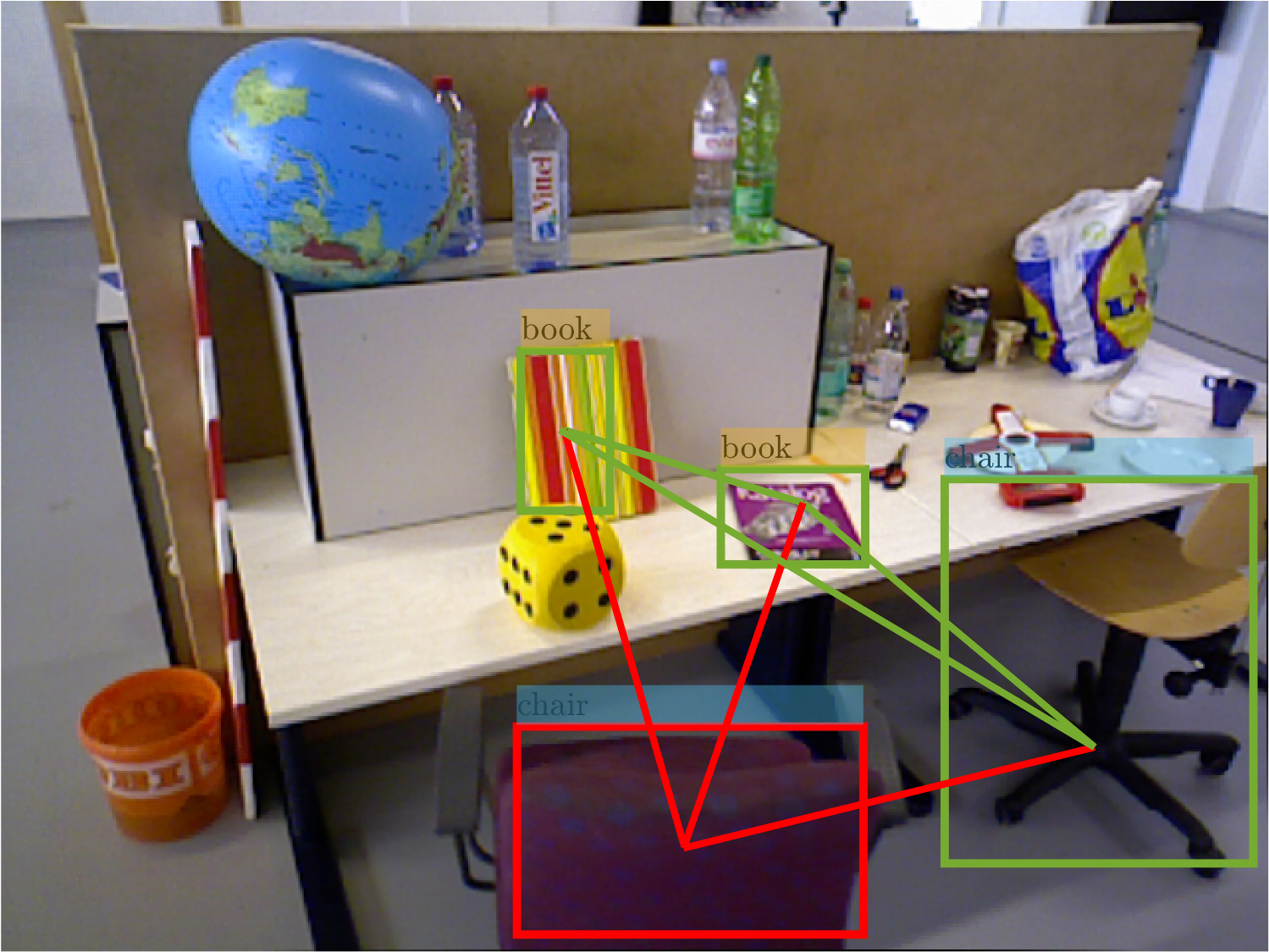}
     \end{subfigure}
     \hfill
     \centering
     \begin{subfigure}[b]{0.493\linewidth}
         \centering
         \includegraphics[width=1\linewidth]{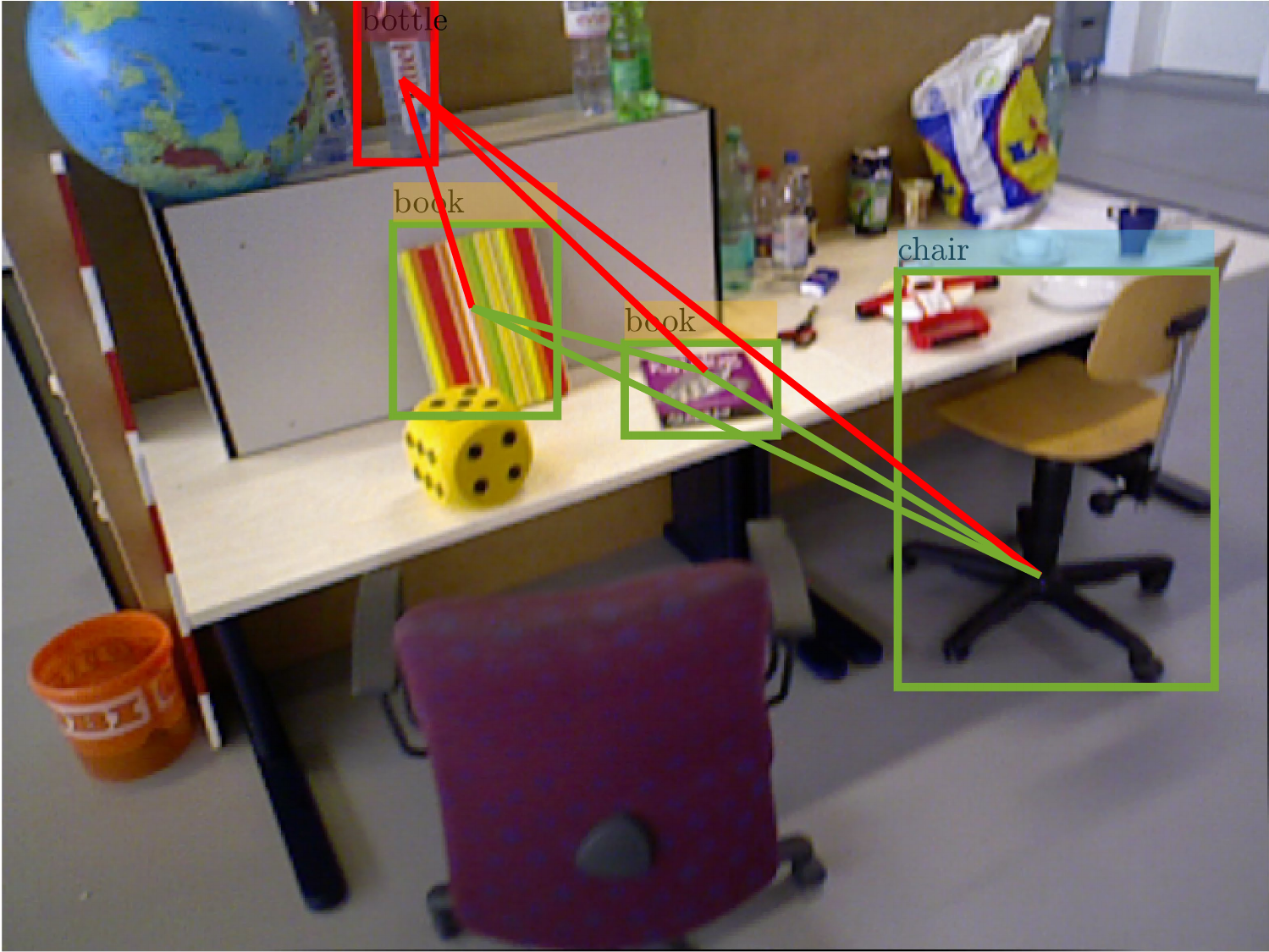}
     \end{subfigure}
        \caption{Covisibility subgraphs on two true loop closure frames (left and right) in  fr3/long\textunderscore office\textunderscore household sequence.  Green and red plots represent matched and mismatched map objects and edges, respectively.} 
        \label{fig:graph in fr2}
\end{figure}

The covisibility subgraphs on the true loop closure frames are shown in Figs. \ref{fig:graph in fr3} and \ref{fig:graph in fr2}. The matched map objects and edges are plotted in green, while the mismatched ones are in red. The two loop closure candidate frames pass the comparison tests for vertices and edges;  thus, our method determined them correctly as true loop closures.
\subsubsection{Ablation Study}
Our method has many 
checks and relies on multiple thresholds. To demonstrate
the contribution of each component to the performance and the roles of the thresholds, we conducted an ablation study with the results shown in Table \ref{tab: ablation}. In the top row, we report the precision and recall of the proposed initial loop candidates. From the top row to the bottom row, we gradually add additional checks as represented by the corresponding thresholds.
We can see that with each additional check, our method is able to identify and reject more false loop closures and increase the precision. Thus each check makes its unique contribution. With all 
checks conducted, our method is able to reject all false loop closures and reach $100\%$ precision for the given data set.

\begin{table}[htbp]
\centering
\caption{Ablation study results}
\begin{tabular}{c c c c c} 
\toprule
 & \multicolumn{2}{ c }{fr2/desk} & \multicolumn{2}{ c }{fr3/long\textunderscore office\textunderscore household}\\
Thresholds & precision & recall & precision & recall\\
\midrule
 None & 11.8\% & 4.3\% & 14.2\% & 5.6\%\\
$+\tau_n$ & 11.9\% & 4.2\% & 14.9\% & 5.3\%\\
$+\tau_{as}$ & 28.9\% & 3.6\% & 53\% & 3.4\%\\
$+\tau_e$  & 29.3\% & 3.6\% & 54.5\% & 3.3\%\\
$+\tau_s$, $+\tau_{epsilon}$ & 100\% & 1.1\% & 100\% & 0.3\%\\
\bottomrule
\end{tabular}
\label{tab: ablation}
\end{table}

\subsubsection{Drift Reduction}
We represent a drift as the root mean square positional error between the estimated trajectory by SLAM and the ground truth trajectory. Benefiting from the coarse-to-fine approach to compute the SIM(3) transformation, the performance of our method in drift reduction is either on par with or better than the baseline methods. Table \ref{tab: Drift on TUM sequences} shows the five-time average of the trajectory errors tested on the two TUM sequences. The entry ``ORB-VO" refers to the result of ORB-SLAM2 with loop closing turned off. Our method obtains the same level of drift reduction in the fr3/long\textunderscore office\textunderscore household sequence and outperforms ORB-SLAM2 and ORB-SLAM3 in the fr2/desk sequence. 
\begin{table}[htbp]
\centering
\caption{Average error on TUM sequences}
\begin{tabular}{c c c } 
\toprule
 & fr2/desk & fr3/long\textunderscore office \textunderscore household \\ 
\midrule
ORB-VO  & 6.0cm & 7.0cm \\
ORB-SLAM2 & 2.0cm & 1.6cm \\
ORB-SLAM3 & 1.5cm & 1.6cm \\
\textbf{SmSLAM+LCD (this paper)} & \textbf{1.4cm} & \textbf{1.6cm} \\ 
\bottomrule
\end{tabular}
\label{tab: Drift on TUM sequences}
\end{table}
\begin{figure}[!t]
  \centering
  \includegraphics[width=0.7\linewidth]{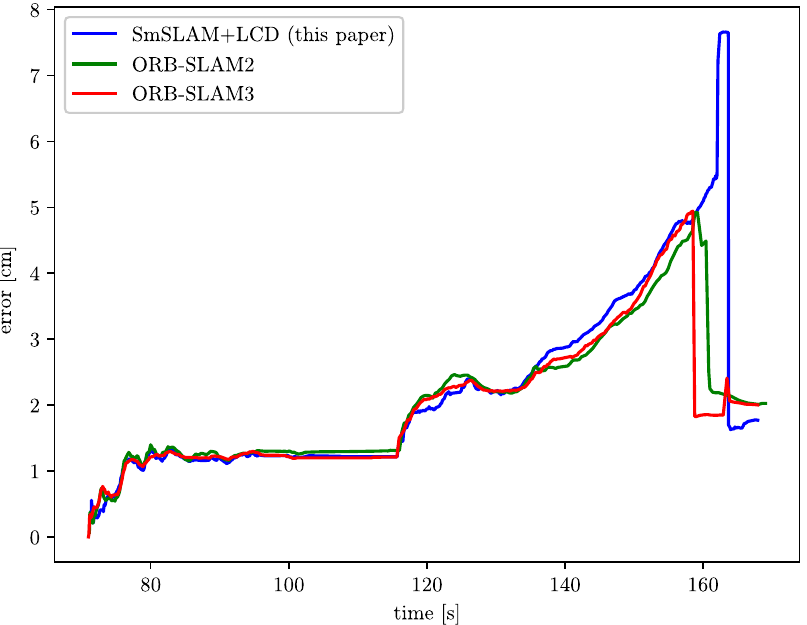}
  \caption{Time-error plots of the fr2/desk sequence.}
  \label{fig:time-error 1}
\end{figure}
\begin{figure}[!t]
  \centering
  \includegraphics[width=0.7\linewidth]{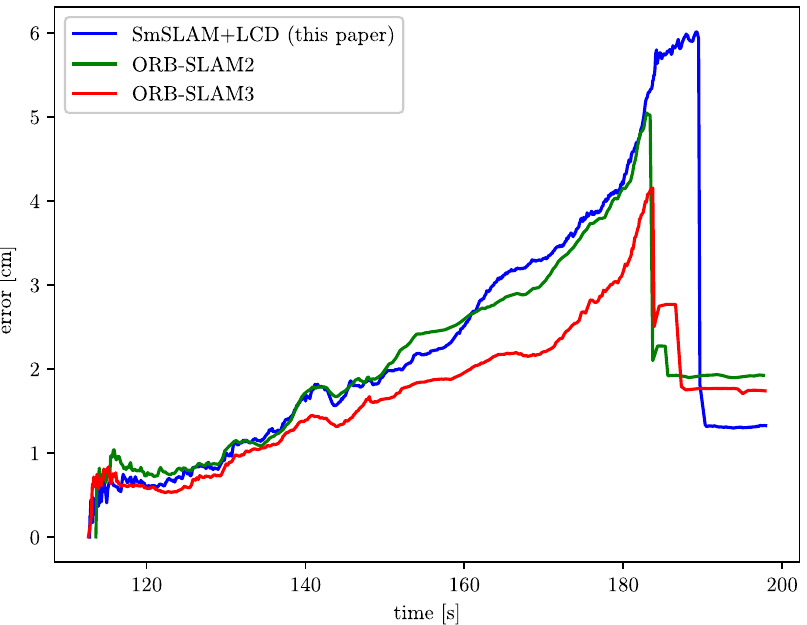}
  \caption{Time-error plots of the fr3/long\textunderscore office\textunderscore household sequence.}
  \label{fig:time-error 2}
\end{figure}

Figs. \ref{fig:time-error 1} and \ref{fig:time-error 2} show the time-error plots of applying our (SmSLAM+LCD) method vs. applying ORB-SLAM2 and ORB-SLAM 3 in the two sequences. The sudden drop in the plot indicates when the loop closure happens. 
Because our method enforces a stricter loop closure check, the loop closure occurs later than ORB-SLAM2 and ORB-SLAM3. For drift reduction, our method shows superior performance. 
\subsection{Results from Virtual Dataset}
The virtual environment is built using Unreal Engine 4 (UE4). UnrealCV \cite{qiu2017unrealcv} is used as the interface between the virtual environment and our system. This environment provides an ideal platform for testing our SmSLAM+LCD method, particularly its ability to distinguish very similar scenes. Our virtual world features a two-story apartment building, modified from the Archviz Interior Rendering sample project built by Epic Games. The second-floor apartment is designed to be similar to the first-floor apartment, though with slightly different furniture and decorations. 

\begin{figure}[!t]
     \centering
     \begin{subfigure}[b]{0.493\linewidth}
         \centering
         \includegraphics[width=1\linewidth]{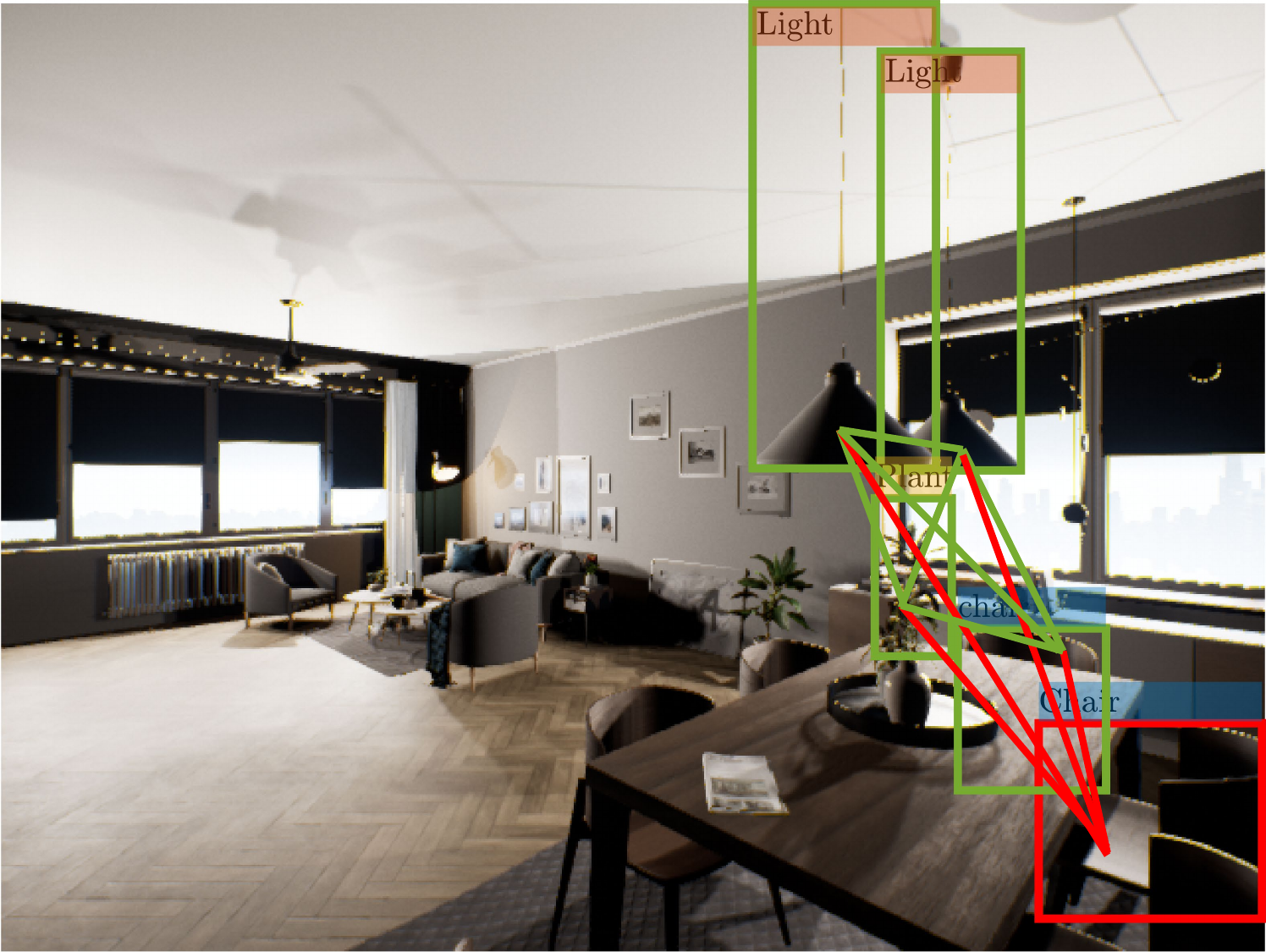}
     \end{subfigure}
     \hfill
     \centering
     \begin{subfigure}[b]{0.493\linewidth}
         \centering
         \includegraphics[width=1\linewidth]{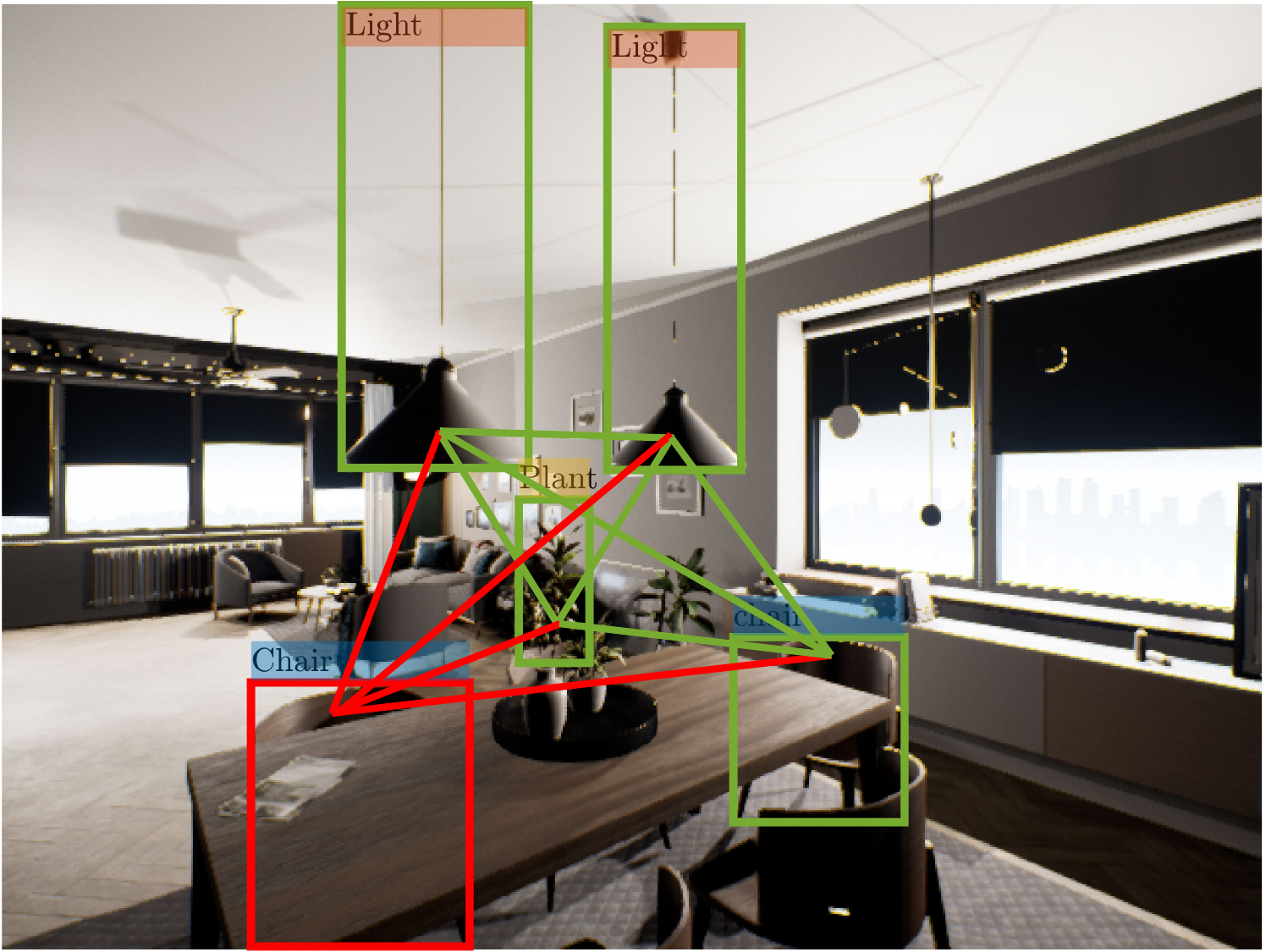}
     \end{subfigure}
        \caption{Covisibility subgraphs on true loop closure frames in the virtual dataset. Green and red plots represent matched and mismatched map objects and edges, respectively.} 
        \label{fig:graph in virtual}
\end{figure}

\begin{figure}[!t]
  \centering
  \includegraphics[width=0.7\linewidth]{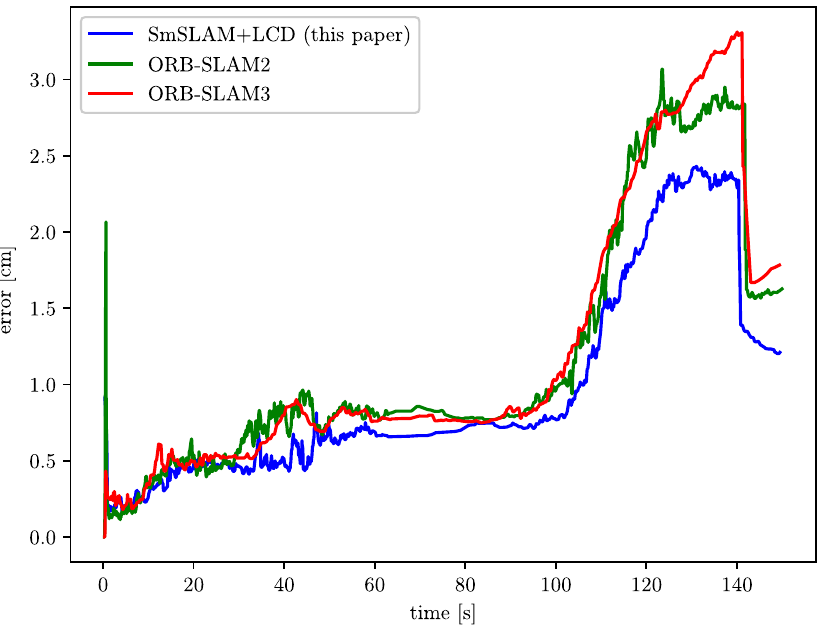}
  \caption{Time-error plots of the virtual dataset.}
  \label{fig:time-error virtual}
\end{figure}

We first tested the loop closure capability in this dataset. A trajectory is designed to traverse the first floor and form a loop. Five-time consecutive tests show that in terms of reducing drifts, our method is able to obtain better performance, benefit from the coarse-to-fine approach to compute the SIM(3) transformation. The average trajectory error with the ORB-VO method is 2.3cm. After the loop is successfully closed, the average drift is reduced to 1.7cm by our method, comparing to 2.1cm by ORB-SLAM2 and 2.1cm by ORB-SLAM3.
The covisibility subgraphs used to detect the true loop closure are shown in Fig. \ref{fig:graph in virtual}. The two covisibility subgraphs are in agreement except for the two chairs marked in red and the edges connected to them. Hence, our method correctly concludes that the two frames form a true loop closure. Fig. \ref{fig:time-error virtual} shows the time-error plots of our method vs. ORB-SLAM2 and ORB-SLAM3. It is clear that our method achieves more significant drift reduction and sooner than ORB-SLAM2 and ORB-SLAM3. 

\begin{figure}[!t]
     \centering
     \begin{subfigure}[b]{0.493\linewidth}
         \centering
         \includegraphics[width=1\linewidth]{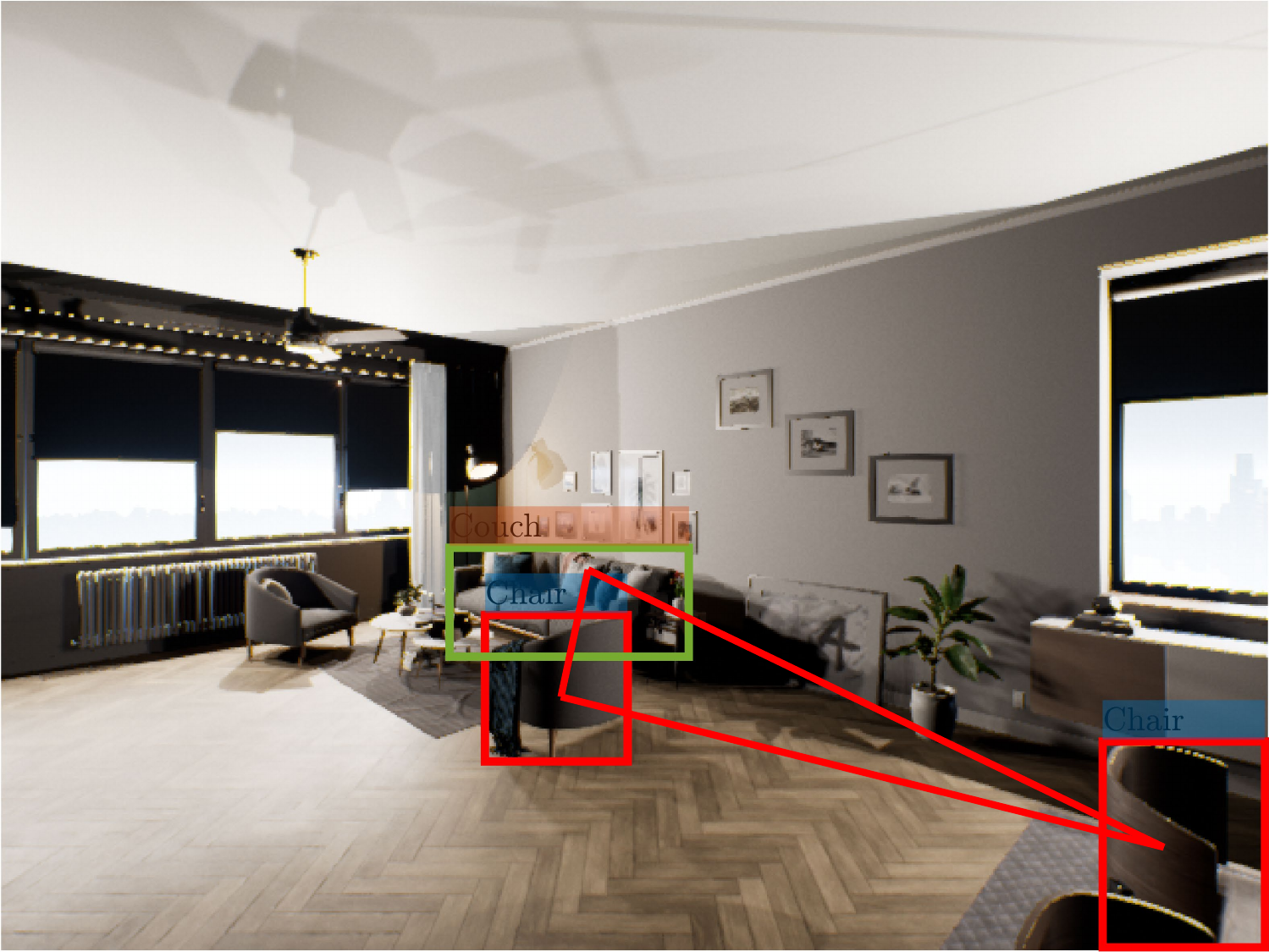}
     \end{subfigure}
     \hfill
     \centering
     \begin{subfigure}[b]{0.493\linewidth}
         \centering
         \includegraphics[width=1\linewidth]{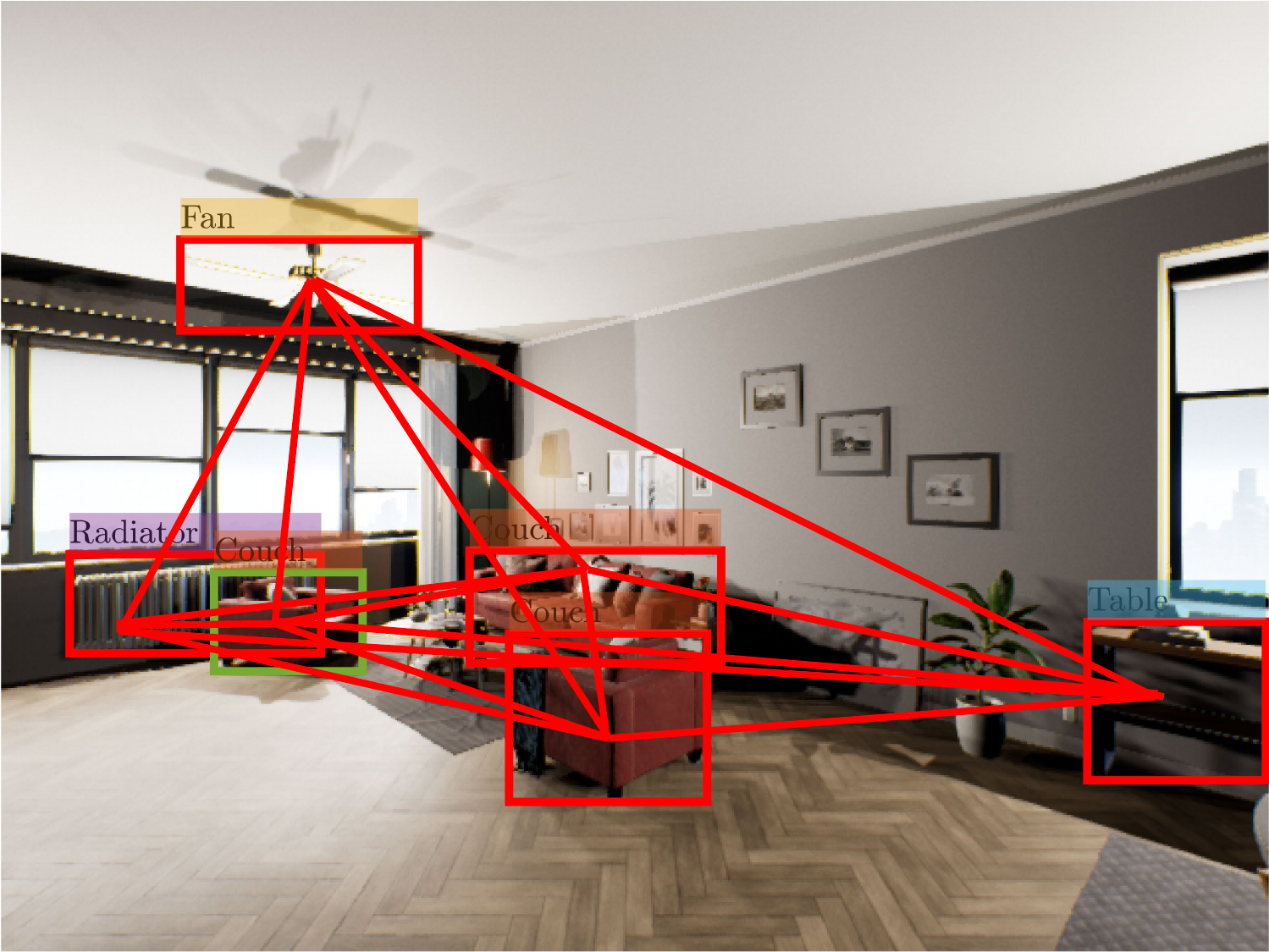}
     \end{subfigure}
        \caption{The false loop closure reported by ORB-SLAM2 and ORB-SLAM3. The left and right frames correspond to the first and second floor apartments. Covisibility subgraphs are also plotted on the two frames.  Green and red plots represent matched and unmatched map objects and edges, respectively.} 
        \label{fig:false loop closure}
\end{figure}

\begin{figure}[!t]
  \centering
  \includegraphics[width=0.7\linewidth]{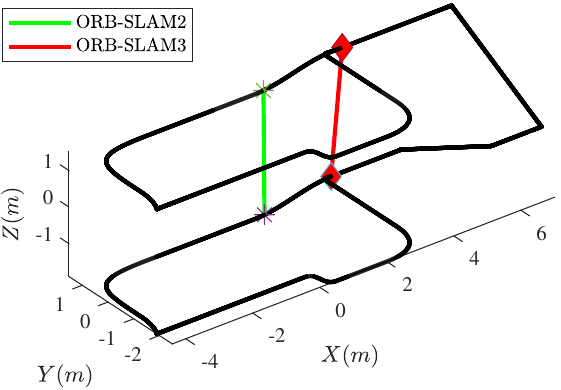}
  \caption{False loop closure visualization on the designed trajectory.}
  \label{fig:false loop closure on tra}
\end{figure}

Another trajectory is designed to traverse the first and the second floors to test the ability to distinguish similar places. Both ORB-SLAM2 and ORB-SLAM3 report a false loop closure between the first-floor apartment image and the second-floor apartment image, as shown in Fig. \ref{fig:false loop closure}. The false loop closures on the designed trajectory are shown in Fig. \ref{fig:false loop closure on tra}, denoted by green and red lines, respectively. In contrast, our method can successfully distinguish the two apartments by comparing the object covisibility subgraphs presented on the two images. An incorrect match occurred between the two couches, represented by green bounding boxes in Fig. \ref{fig:false loop closure}, but all other edges and vertices of the two covisibility subgraphs are not matched. Hence our method correctly concludes that the two images were taken at two different places.  

\subsection{Results from Real-world Dataset}
\begin{figure}[!t]
     \centering
     \begin{subfigure}[b]{0.493\linewidth}
         \centering
         \includegraphics[width=1\linewidth]{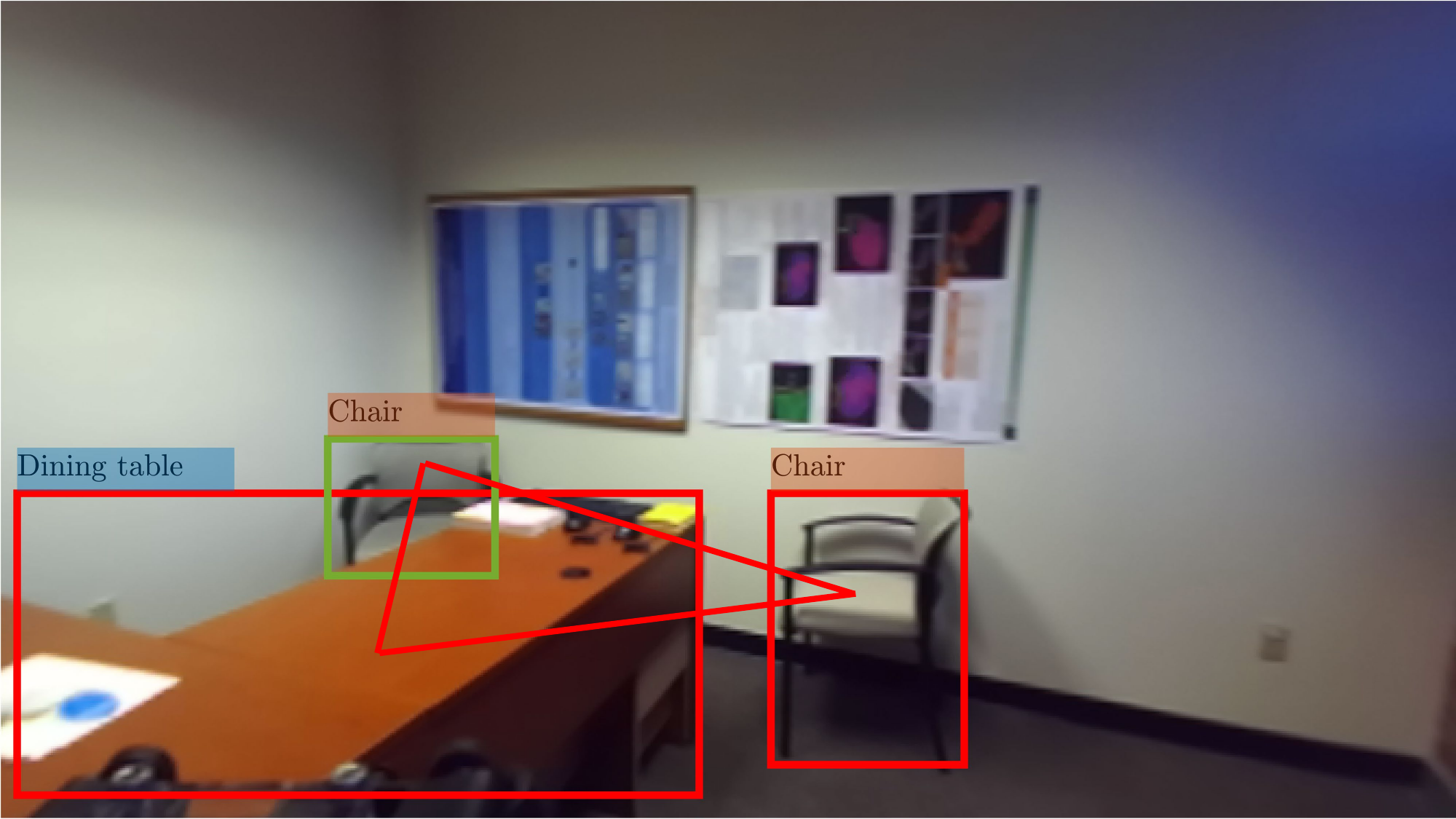}
     \end{subfigure}
     \hfill
     \centering
     \begin{subfigure}[b]{0.493\linewidth}
         \centering
         \includegraphics[width=1\linewidth]{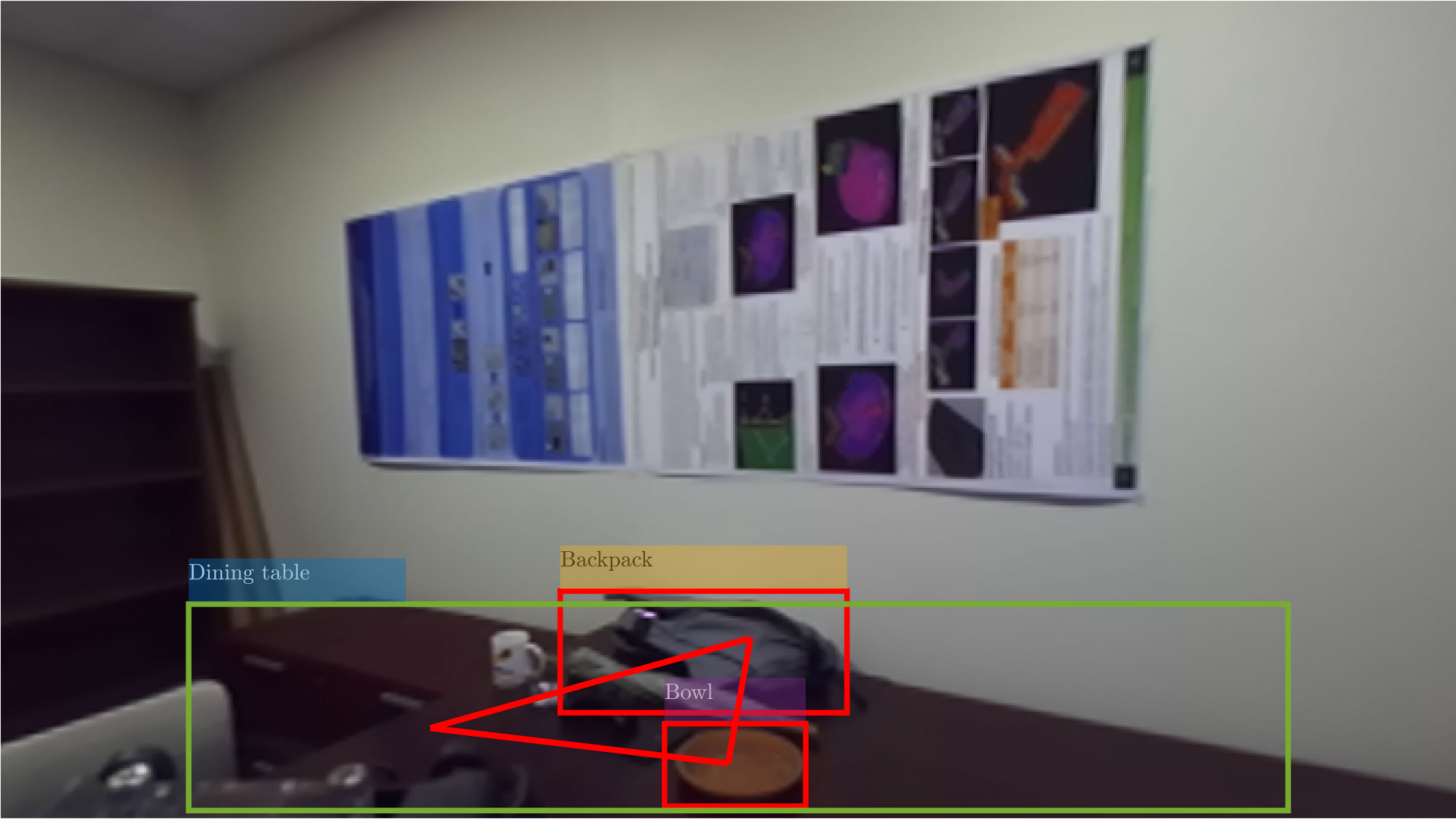}
     \end{subfigure}
        \caption{The false loop closure reported by ORB-SLAM2 and ORB-SLAM3 but not by our method. The left and right frames correspond to the first and second offices, respectively. Covisibility subgraphs are also plotted on the two frames.  Green and red plots represent matched and unmatched map objects and edges, respectively.} 
        \label{fig:real false loop closure}
\end{figure}

Finally, we tested our algorithm in a real-world dataset. The dataset is constructed by a human operator carrying a ZED camera and traversing two adjacent offices. Posters with the same content are placed on the walls of the two offices to create two similar scenes. ORB-SLAM2 and ORB-SLAM3 report false loop closure when tested in this dataset. In contrast, our algorithm can successfully distinguish the two offices apart by comparing the covisibility subgraphs presented on loop candidate frames, as shown in Fig. \ref{fig:real false loop closure}. The chair in the first office is incorrectly matched with the dining table in the second office by our method, possibly because the detection of the dining table contains portions of a similar chair. Nevertheless, our method successfully distinguished other map objects of the two covisibility subgraphs and ruled out this false loop closure. 

\begin{figure}[!t]
  \centering
  \includegraphics[width=0.69\linewidth]{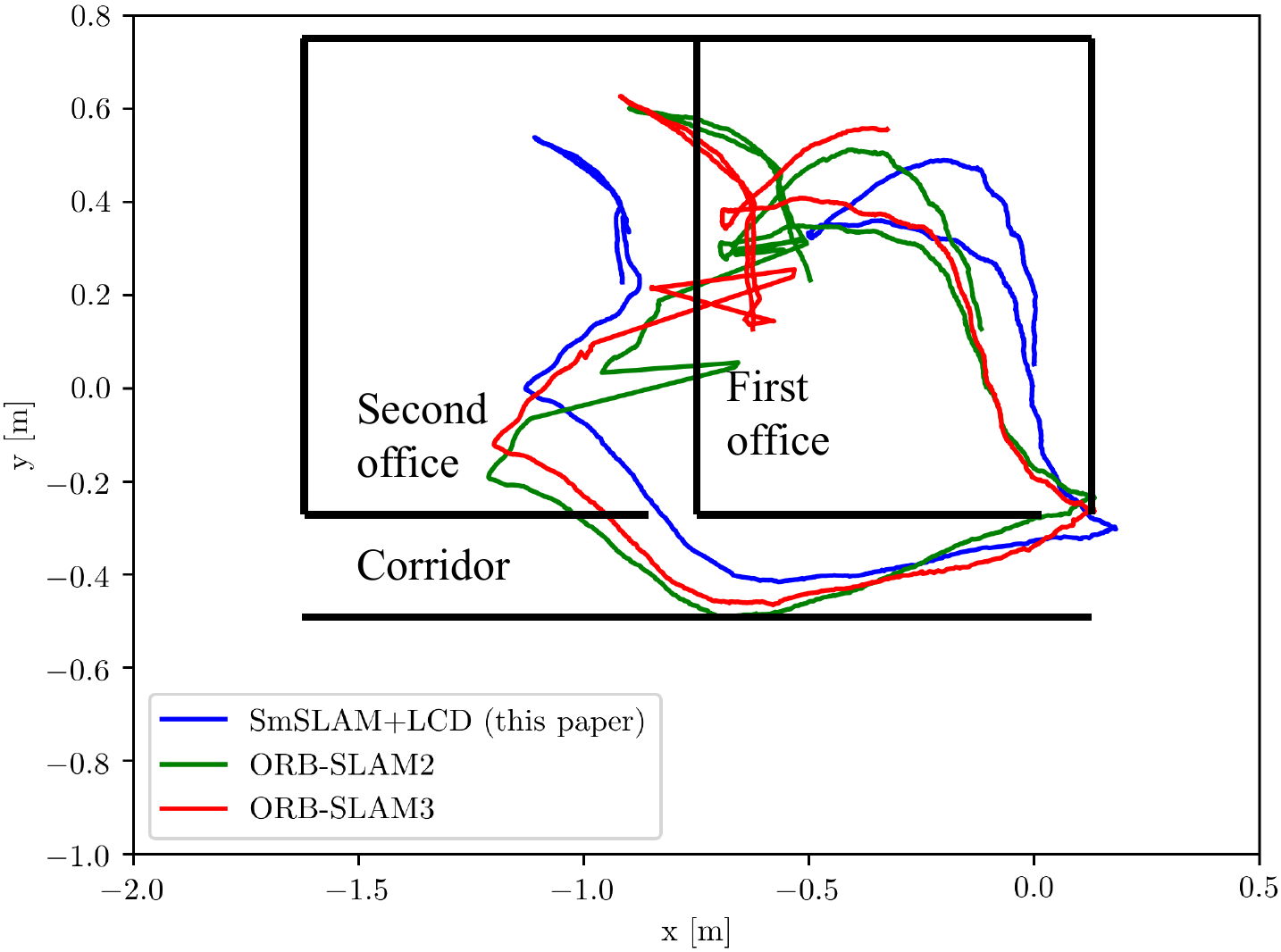}
  \caption{Estimated sensor poses in the real-world dataset.}
  \label{fig:trajectory real}
\end{figure}

The estimated sensor poses of our method, ORB-SLAM2, and ORB-SLAM3 in the real-world dataset are plotted in Fig. \ref{fig:trajectory real}. 
Black lines in the image mark the rough layout of the real-world environment. From Fig. \ref{fig:trajectory real}, we can see that because ORB-SLAM2 and ORB-SLAM3 performed a false-positive loop closure, the estimated trajectories are severely distorted and cut through walls. In contrast, our method successfully rejected the false loop closure and maintained more accurate tracking.
 
%% file: sections/conclusion.tex
In this paper, we have introduced SmSLAM+LCD, a semantic loop closure method well integrated into a semantic SLAM system of our prior work \cite{qian2020semantic}. Our method leverages high-level semantic and low-level geometric information from the SLAM system for loop detection and drift correction. In particular, SmSLAM+LCD builds an object covisibility graph upon the mapped 3D semantic objects and constantly updates it with the latest observation. When performing loop detection, our method checks the loop candidates proposed based on low-level geometric features by comparing the object covisibility subgraphs associated with the loop candidate frames. To correct the accumulated drift, our method includes a coarse-to-fine approach to compute the SIM(3) transformation between loop closure frames. 

We tested SmSLAM+LCD using the TUM RGB-D dataset and our own virtual and real-world datasets. The experimental results show that SmSLAM+LCD can achieve more accurate drift correction after detecting a loop closure than ORB-SLAM2 and ORB-SLAM3 and distinguish similar scenes to avoid false-positive loop closures reported by ORB-SLAM2 and ORB-SLAM3. 

In this paper, we assume  
that every object contributes equally to the detection of a true loop closure or the rejection of a false loop closure. This could be modified to treat static and moving objects differently, which we plan to do next. 
We will also investigate using weights in the covisibility graphs to improve loop detection in dynamic environments.
